\documentclass{article}
\usepackage{spconf,amsmath,amssymb,graphicx}
\usepackage{xcolor}         
\usepackage{bm}
\usepackage{algorithm}
\usepackage[noend]{algpseudocode}
\usepackage{cleveref}

\title{Diversity-aware Buffer for Coping with Temporally Correlated Data Streams in Online Test-time Adaptation}
\name{Mario D\"obler, Florian Marencke, Robert A. Marsden, and Bin Yang}
\address{University of Stuttgart\\
{\tt\small  \{mario.doebler, robert.marsden, bin.yang\}@iss.uni-stuttgart.de}}
\begin{document}
%
\maketitle
\begin{abstract}
Since distribution shifts are likely to occur after a model's deployment and can drastically decrease the model's performance, online test-time adaptation (TTA) continues to update the model during test-time, leveraging the current test data. In real-world scenarios, test data streams are not always independent and identically distributed (i.i.d.). Instead, they are frequently temporally correlated, making them non-i.i.d. Many existing methods struggle to cope with this scenario. In response, we propose a diversity-aware and category-balanced buffer that can simulate an i.i.d. data stream, even in non-i.i.d. scenarios. Combined with a diversity and entropy-weighted entropy loss, we show that a stable adaptation is possible on a wide range of corruptions and natural domain shifts, based on ImageNet. We achieve state-of-the-art results on most considered benchmarks.
\end{abstract}
\begin{keywords}
test-time adaptation, computer vision
\end{keywords}
\section{Introduction}
\label{sec:intro}
Deep neural networks achieve remarkable performance, as long as training and test data originate from the same distribution. However, in real-world applications, conditions can change during test-time, leading to a decline in performance of the deployed model. To address potential domain shifts, domain generalization aims to improve the robustness and generalization of the model directly during training. Due to the broad range of data shifts \cite{quinonero2008dataset} which are typically unknown during training \cite{mintun2021interaction}, the effectiveness of these approaches is limited. To address this issue, online test-time adaptation (TTA) has emerged. In online TTA, the model is adapted directly during test-time using an unsupervised loss function and the available test sample(s) at time step $t$, which provide insights into the current distribution.

Although TENT \cite{wang2021tent} has demonstrated success in adapting to i.i.d. data, recent research on TTA has identified more challenging scenarios where methods solely based on self-training, such as TENT, often fail \cite{marsden2022gradual, niu2023towards, gong2022note, yuan2023robust, boudiaf2022parameter}. In particular, coping with temporally correlated data streams remains an open problem. One recent line of work \cite{gong2022note, yuan2023robust} tackles this problem by simulating a uniform data stream through the introduction of a buffer. Instead of utilizing the current test batch $\bm{x}_t$ to update the model, the incoming samples are stored in a buffer. Different criteria are employed to maintain a uniform class distribution within the buffer. Due to memory limits, existing samples in the buffer have to be removed. Finally, a batch is sampled from the buffer to update the model. NOTE \cite{gong2022note} proposes Prediction-Balanced Reservoir Sampling (PBRS) that combines time-uniform and prediction-uniform sampling. RoTTA \cite{yuan2023robust} introduces Category-balenced Sampling with Timeliness and Uncertainty (CSTU) that promotes recent and certain samples in the buffer. Alongside PBRS and CSTU, our proposed Diversity-aware Buffer (DAB) employs a category-balanced buffer, but reduces the redundancy within the buffer by only adding diverse samples to the buffer. Moreover, we only update the model when enough samples from different classes have been replaced, reducing the correlation among consecutive updates. While it is reasonable to promote certain samples within the buffer as in RoTTA, we follow recent ideas of diversity and certainty weighting \cite{niu2022efficient, marsden2024universal}. Here, the idea is to scale the self-training loss by certainty and diversity-based weights, resulting in a larger contribution of diverse and certain samples. To further ensure a stable adaptation, we leverage weight ensembling from \cite{marsden2024universal}, where after each update the current model weights are averaged with a small percentage of the source model. This ensures that the model cannot drift too far apart from the source model.

In cases of temporally correlated data, another challenge arises in estimating reliable batch normalization (BN) statistics. For robust statistics, NOTE \cite{gong2022note} introduced an instance-aware batch normalization and RoTTA \cite{yuan2023robust} employs a robust batch normalization variant. Alternatively, one can leverage normalization layers like group or layer normalization, which do not require a batch of data to estimate the statistic and are thus better suited \cite{TTT, niu2023towards}.

Our contributions are as follows. We propose a simple but novel category-balanced buffer that only stores diverse samples to reduce the redundancy. To reduce the correlation between consecutive updates, we only update the model when enough samples from different classes have been replaced. Additionally, inspired by recent work, we scale the self-training loss with certainty and diversity-based weights and employ weight ensembling. We empirically demonstrate the effectiveness of our proposed method DAB on a wide range of domain shifts based on ImageNet.

\section{Methodology}
Let $\bm{\theta}_0$ denote the weights of a deep neural network pre-trained on labeled source data $(\mathcal{X}, \mathcal{Y})$. Typically, the network will perform well on data originating from the same domain. However, when faced with data from different domains, performance is likely to deteriorate. To ensure that the networks' performance remains high during inference, online test-time adaptation continues to update the model after deployment using an unsupervised loss function, such as the entropy, and the currently available test data $\bm{x}_t$ at time step $t$. For a successful model adaptation, minimizing the entropy requires batches of independent and identically distributed (i.i.d.) data. In practice, this assumption is often violated, a model can encounter multiple domains and temporally correlated data, denoted as Practical TTA in \cite{yuan2023robust}. Further, for temporally correlated data, a reliable estimation of the BN statistics, which are commonly used by recent TTA methods \cite{wang2021tent, wang2022continual, dobler2023robust}, is not possible.

Therefore, we rely on architectures that do not employ batch normalization layers and introduce a diversity-aware and category-balanced buffer to effectively adapt to both i.i.d. and correlated data streams. To reduce redundancy in the buffer, only incoming samples are stored in the buffer, if they fulfill our diversity criteria as described in \Cref{sec:diversity_buffer}. In case the buffer is at capacity, the oldest sample from the majority class is removed to maintain a category-balanced buffer. To reduce the correlation between consecutive updates, we only update the model when enough samples from different classes have been replaced. Once this is the case, a batch is uniformly sampled from the buffer to minimize a weighted entropy loss, as introduced in \Cref{sec:loss_weighting}. To ensure efficiency, only the network's normalization parameters are updated. For an overall stable adaptation, we employ weight ensembling from \cite{marsden2024universal}. After each update the weights of the initial source model $\bm{\theta}_0$ and the weights of the current model $\bm{\theta}_t$ at time step $t$ are averaged using an exponential moving average $
    \bm{\theta}_{t+1} = \alpha \, \bm{\theta}_t + (1-\alpha)\bm{\theta}_0$ ,
where $\alpha$ is a momentum term. It serves as a corrective measure, capable of rectifying suboptimal adaptations over time, by continually incorporating a small percentage of the source weights.

\subsection{Diversity-aware Buffer}
\label{sec:diversity_buffer}

\begin{algorithm*}[t]
\caption{Diversity-aware Buffer}\label{alg:dab}
    \begin{algorithmic}[1]
    
    \Require {current test batch $\bm{x}_t$ with size $N$, buffer $B$ of capacity $M$
    }
    \State $B \leftarrow \varnothing$; and $n[c] \leftarrow 0$ for $c \in \mathcal{Y}$
    \For {$t\in\{1, \cdots, T\}$}
        \State Compute $\bm{w}_{\mathrm{div}, t}$ \textcolor{gray}{\quad// according to \Cref{eq:diversity_weights}}
        \For {$i\in\{1, \cdots, N\}$}
            \If {$w_{\mathrm{div},{ti}}>\mathrm{mean}(\bm{w}_{\mathrm{div}, t})$} \textcolor{gray}{\quad// only add diverse samples to buffer}
                \State $n[\hat{c}_{ti}] \leftarrow n[\hat{c}_{ti}] + 1 \quad \mathrm{with} \quad \hat{c}_{ti}=\arg\max_{c}{\hat{\bm{y}}_{tic}}$ \textcolor{gray}{\quad// increase the number of samples encountered for the class}
                \State $b[c] \leftarrow |\{ (\bm{x}, y) \in B | y = c \}|$ for $c \in \mathcal{Y}$ \textcolor{gray}{\quad// count instances per class in buffer}
                \If {$|B|\ < M$} \textcolor{gray}{\quad// if buffer is not full}
                    \State Add $(\bm{x}_{ti}, \hat{c}_{ti})$ to $B$
                \Else
                    \State $C^*\gets \arg\max_{c \in \mathcal{Y}} b[c]$ \textcolor{gray}{\quad// get majority class(es)}
                        \State Pick oldest $B[j] := (\bm{x}_j, \hat{c}_{j})$ where $\hat{c}_{j} \in C^*$ 
                        \State $B[j] \leftarrow (\bm{x}_{ti}, \hat{c}_{ti})$ \textcolor{gray}{\quad// replace it with a new sample}
                        \State $n[\hat{c}_{j}] \leftarrow n[\hat{c}_{j}] - 1$ \textcolor{gray}{\quad// decrease the class counter correspondingly}
                \EndIf
            \EndIf
        \EndFor
    \EndFor
    \end{algorithmic}
\end{algorithm*}

Since temporally correlated distributions lead to an undesirable class bias, adapting a model with consecutive test samples, especially when they belong to the same class, negatively impacts the optimization objective, such as the entropy. To combat this imbalance, we propose a diversity-aware and category-balanced buffer $B$ of capacity $M$. Now, sampling a batch from the buffer should result in i.i.d. data, even from non-i.i.d. data streams, enabling a stable model adaptation.

Since samples can be redundant within a batch, in contrast to previous work, we only want to store samples in the buffer that are diverse with respect to the model's output distribution. We begin by tracking the recent tendency of a model's softmax output $\bm{\hat{y}}_{ti}$ with an exponential moving average $\bm{\bar{y}}_{t+1}=\beta\, \bm{\bar{y}}_{t} + \frac{(1-\beta)}{N} \sum_i^{N}\bm{\hat{y}}_{ti}$, setting $\beta=0.9$. To determine a diversity weight for each test sample $\bm{x}_{ti}$, the cosine similarity between the current model output $\bm{\hat{y}}_{ti}$ and the tendency of the recent outputs $\bm{\bar{y}}_{t}$ is computed as follows
\begin{equation}
\label{eq:diversity_weights}
    w_{\mathrm{div},{ti}}=1-\frac{\bm{\hat{y}}_{ti}^\mathrm{T} \, \bm{\bar{y}}_{t}}{\Vert\bm{\hat{y}}_{ti}\Vert \, \Vert\bm{\bar{y}}_{t}\Vert}.
\end{equation}
Samples that deviate from the recent output tendency $\bm{\bar{y}}_{t}$ receive a large weight, while samples that are similar, receive a small weight. This ensures that, e.g., when a class momentarily dominates the test stream, samples from other classes are favored for being added to the buffer.

In particular, only samples that fulfill the diversity criterion $w_{\mathrm{div},{ti}}>\mathrm{mean}(\bm{w}_{\mathrm{div}, t})$ are stored in the buffer. In case the buffer is at capacity, the oldest sample from the majority class is replaced. This results in an up-to-date and category-balanced buffer. To reduce the correlation between consecutive updates, we employ the following strategy: Once $N/4$ samples from different classes have been replaced in the buffer, a batch $\tilde{\bm{x}}$ of size $N$ is uniformly sampled from the buffer. The batch is used to minimize a weighted entropy loss, as described in \Cref{sec:loss_weighting}. Details about the diversity-aware buffer are presented in \Cref{alg:dab}.

\subsection{Diversity and Entropy-based Loss Weighting}
\label{sec:loss_weighting}
A common approach for online test-time adaptation involves using the entropy as a self-training loss. However, not all samples are equally reliable. We draw inspiration from recent studies by \cite{niu2022efficient, marsden2024universal} and introduce a diversity and entropy-based scaling factor $w$ for the entropy
\begin{equation}
    \mathcal{L}_\mathrm{ENT}(\bm{\hat{y}}_{i})=-\sum_{c} w_{i}\, \hat{y}_{ic}\log\hat{y}_{ic}.
\end{equation}
To ensure efficiency during test-time, we only update the network's normalization parameters and freeze all others.

In particular, for the batch $\tilde{\bm{x}}$ sampled from the buffer, we use the same diversity scheme as in \Cref{sec:diversity_buffer}. We track a separate recent tendency of a model's prediction $\bm{\hat{y}}_{ti}$ based on the outputs of the sampled batches from the buffer. The diversity weights for $\tilde{\bm{x}}$ are then calculated using \Cref{eq:diversity_weights}. To remove dependencies on model-specific factors or data characteristics, we normalize the diversity weights to be in unit range. To pull apart diverse and non-diverse samples, we take the exponential of the diversity weights. Further, to promote reliable samples receiving a large certainty weight, we utilize the entropy
\begin{equation}
\frac{1}{w_{\mathrm{cert},i}}=\frac{H(\bm{\hat{y}}_{i})}{H_\mathrm{max}}=\frac{\sum_c \hat{y}_{ic}\log \hat{y}_{ic}}{\sum_c \frac{1}{|C|}\log \frac{1}{|C|}},
\end{equation}
which is normalized by the maximum entropy of a uniform prediction. To limit the certainty weight range, we clamp each certainty weight $w_{\mathrm{cert},i}$ to be in range $[1, 10]$. For the combined weights $\bm{w}$, we use the element-wise multiplication of certainty weights $\bm{w}_{\mathrm{cert}}$ and diversity weights $\bm{w}_{\mathrm{div}}$
\begin{equation}
    \bm{w}=\bm{w}_{\mathrm{cert}}\exp(\bm{w}_{\mathrm{div}}).
\end{equation}

\section{Experiments}
\subsubsection{Datasets}
We consider the corruption benchmark ImageNet-C \cite{hendrycks2019benchmarking}, including 15 types with 5 severity levels. For natural domain shifts, we consider ImageNet-R \cite{hendrycks2021many}, ImageNet-Sketch \cite{wang2019learning}, as well as ImageNet-D109. While ImageNet-R contains 30,000 examples depicting different renditions of 200 IN classes, ImageNet-Sketch contains 50 sketches for each of the 1,000 IN classes. ImageNet-D109 \cite{marsden2024universal} is based on DomainNet \cite{peng2019moment} and contains 5 domain shifts (clipart, infograph, painting, real, sketch) with varying domain lengths. While for ImageNet-C and ImageNet-Sketch the classes are uniformly distributed, this is not the case for ImageNet-R and ImageNet-D109. For ImageNet-R it varies from 51 to 430 samples per class, for ImageNet-D109, it depends on the domain, e.g., for clipart it varies from 12 to 469.

\subsubsection{Considered settings}
All experiments are performed in the online TTA setting, where the predictions are evaluated immediately. To assess the performance of each method, we consider the continual and correlated settings. In case of the \textit{continual} benchmark \cite{wang2022continual}, the model is adapted to a sequence of $K$ different domains $\mathcal{D}$ without knowing when a domain shift occurs, i.e. $[\mathcal{D}_1, \mathcal{D}_2, \dots, \mathcal{D}_K]$. For ImageNet-C, the domain sequence comprises 15~corruptions, each encountered at the highest severity level~5. For ImageNet-R and ImageNet-Sketch there exists only a single domain and for ImageNet-D109 the domains are encountered in alphabetical order. In the \textit{correlated} setting the domains are also encountered sequentially. However, the samples of each domain is sorted by the class label rather than randomly shuffled, resulting in class imbalanced batches. While for ImageNet-R, ImageNet-Sketch, and ImageNet-D109 the samples are identical for the \textit{continual} and \textit{correlated} setting, for ImageNet-C this is not the case, due to the protocol from \cite{wang2022continual}. In the \textit{continual} setting, the sequence only consists of 5,000 samples per domain. In the \textit{correlated} setting all samples, namely 50,000 samples per domain, are used.

\subsubsection{Implementation details}
For all datasets a source pre-trained VisionTransformer \cite{dosovitskiy2020image} in its base version with an input patch size of $16\times16$ (Vit-b-16), is used. We follow the implementation of \cite{wang2021tent}, using the same hyperparameters. For all datasets, a batch size $N$ of $64$ is employed. As an optimizer SGD with a learning rate of $2.5e-4$ and a momentum of $0.9$ is used. For weight ensembling we use a momentum of $\alpha=0.99$.

\subsubsection{Baselines}
We compare our approach to other source-free TTA methods that also use an arbitrary off-the-shelf pre-trained model. In particular, we compare to TENT non-episodic \cite{wang2021tent}, CoTTA \cite{wang2022continual}, AdaContrast \cite{chen2022contrastive}, EATA \cite{niu2022efficient}, SAR \cite{niu2023towards}, RoTTA \cite{yuan2023robust}, and NOTE \cite{gong2022note}. Both RoTTA and NOTE employ a buffer for dealing with temporally correlated data streams. In addition, we report the performance of the non-adapted model (source). As a metric, we consider the error rate.

\subsection{Results}
\textbf{Continual TTA}
\begin{table}[t]
\renewcommand{\arraystretch}{1.2}
\centering
\caption{Online classification error rate~(\%) in the \textit{continual} TTA setting, averaged over 5 runs. Results worse than the source performance are highlighted in red.} \label{tab:continual}
\tabcolsep4pt
\begin{tabular}{l|c|cccc}\hline
Method & \rotatebox[origin=c]{90}{ Buffer size } & \rotatebox[origin=c]{70}{ImageNet-C} & \rotatebox[origin=c]{70}{ImageNet-R} & \rotatebox[origin=c]{70}{ImageNet-Sk.} & \rotatebox[origin=c]{70}{ImageNet-D109}  \\
\hline
Source & -  & 60.2 & 56.0 & 70.6 & 53.6  \\
\hline
TENT        & -  & 54.5 & 53.3 & 70.5 & \textcolor{red}{84.0}  \\
CoTTA       & -  & \textcolor{red}{77.0} & \textcolor{red}{69.6} & \textcolor{red}{95.5} & \textcolor{red}{73.4}  \\
AdaContrast & -  & 57.0 & 54.2 & 68.3 & 49.7  \\
EATA        & -  & 49.8 & 49.0 & \textbf{59.7} & 47.4  \\
SAR         & -  & 51.7 & 48.6 & 70.6 & \textcolor{red}{57.4} \\
RoTTA       & 64 & 58.3 & 54.4 & 69.0 & 51.2  \\
NOTE        & 64 & 54.2 & 51.8 & 63.5 & 49.5  \\
\hline
DAB (ours)  & 64 & \textbf{47.4} & 47.8 & 60.9 & 47.1  \\
DAB (ours)  & 256 & 48.2 & \textbf{47.5} & 60.9 & \textbf{46.8} \\
\hline
\end{tabular}
\vskip -14pt
\end{table}

\Cref{tab:continual} shows the results for online continual TTA. In the continual setting, methods solely based on self-training, such as TENT, show a stable adaptation for datasets with moderate lengths. When the sequence is too long and contains multiple domain shifts, TENT is likely to collapse at some point \cite{marsden2024universal}, as demonstrated by the performance on ImageNet-D109. CoTTA which has been optimized for different model architectures do not show to be model-agnostic and show an unstable model adaptation for all considered benchmarks. AdaContrast and the baselines RoTTA and NOTE that employ a buffer can all improve upon the source performance. Our method DAB, significantly improves upon the source performance and is on average $3.8\%$ better than the second best method that uses a buffer: NOTE. EATA shows the best performance among the variants without a buffer, but our method DAB still outperforms EATA on three out of the four continual benchmarks.

\textbf{Correlated TTA}
\begin{table}[t]
\renewcommand{\arraystretch}{1.2}
\centering
\caption{Online classification error rate~(\%) in the \textit{correlated} TTA setting, averaged over 5 runs. Results worse than the source performance are highlighted in red.} \label{tab:correlated}
\tabcolsep4pt
\begin{tabular}{l|c|cccc}\hline
Method & \rotatebox[origin=c]{90}{ Buffer size } & \rotatebox[origin=c]{70}{ImageNet-C} & \rotatebox[origin=c]{70}{ImageNet-R} & \rotatebox[origin=c]{70}{ImageNet-Sk.} & \rotatebox[origin=c]{70}{ImageNet-D109}  \\
\hline
Source & -  & 60.2 & 56.0 & 70.6 & 53.6  \\
\hline
TENT        & -  & \textcolor{red}{80.6} & 53.4 & 66.7 & \textcolor{red}{84.3}  \\
CoTTA       & -  & \textcolor{red}{98.8} & \textcolor{red}{81.0} & \textcolor{red}{95.5} & \textcolor{red}{93.1}  \\
AdaContrast & -  & \textcolor{red}{87.4} & \textcolor{red}{62.1} & \textcolor{red}{72.3} & \textcolor{red}{56.7}  \\
EATA        & -  & \textcolor{red}{76.2} & 53.6 & 63.7 & \textcolor{red}{57.4}  \\
SAR         & -  & 53.9 & \textbf{49.9} & \textcolor{red}{74.6} & \textcolor{red}{58.7}  \\ 
RoTTA       & 64 & \textcolor{red}{65.1} & 55.8 & 70.1 & \textcolor{red}{53.8}  \\
\hline
NOTE        & 64 & \textcolor{red}{89.4} & 53.4 & 66.7 & 54.1  \\
NOTE        & 256 & \textcolor{red}{82.6} & 52.8 & 65.1 & 51.4  \\
NOTE        & 1024 & \textcolor{red}{68.8} & 52.5 & 64.5 & 50.7  \\
NOTE        & 4096 & \textcolor{red}{62.9} & 52.6 & 64.2 & 50.7 \\ 
\hline
DAB (ours)  & 64 & \textcolor{red}{69.1} & 56.0 & \textcolor{red}{74.4} & \textcolor{red}{57.0} \\
DAB (ours)  & 256 & \textcolor{red}{62.1} & 53.5 & 69.5 & 50.5  \\
DAB (ours)  & 1024 & 55.2 & 50.9 & 65.3 & \textbf{49.0}  \\
DAB (ours)  & 4096 & \textbf{49.6} & 50.6 & \textbf{63.7} & 49.1  \\ 
\hline
\end{tabular}
\vskip -14pt
\end{table}

\Cref{tab:correlated} shows the results for online correlated TTA. In the correlated setting, adaptation is much more difficult, as shown by the performance of the methods that do not employ a buffer. Even RoTTA that uses a buffer of size $64$ cannot improve upon the source performance on average. Both NOTE (with the exception of ImageNet-C) and our method DAB show a stable adaptation as long as the buffer size is large enough. For all considered benchmarks, except ImageNet-C, a buffer size of $256$ is sufficient. For ImageNet-C much larger buffer sizes lead to significant improvements. While buffer-free SAR shows a stable adaptation for ImageNet-C and ImageNet-R, the performance drops on ImageNet-Sketch and ImageNet-D109, even though the method was proposed for such scenarios. This highlights the requirement of a buffer in non-i.i.d. settings. Our method DAB is the only method that can improve upon the source performance for all benchmarks, closing the performance gap between the continual and correlated setting. For a buffer size of $4096$, we reduce the error by $4.3\%$ compared to NOTE and $6\%$ compared to the best buffer-free method: SAR.

\textbf{Influence of buffer size}
In \Cref{tab:correlated} we additionally ablate the buffer size for NOTE and our method DAB. We find that the optimal buffer size depends on the considered benchmark. For ImageNet-R, ImageNet-Sketch, and ImageNet-D109, a buffer size in the range $[1024, 4096]$ shows to be a good choice. For DAB a buffer size of $4096$ leads to further significant gains on ImageNet-C.

\section{Conclusion}
In this work we proposed a diversity-aware and category-balanced buffer to cope with temporally correlated data streams. For a stable adaptation, we additionally introduced a diversity and entropy-weighted entropy loss with weight ensembling. We set state-of-the art results on various benchmarks based on ImageNet.

\newpage
\bibliographystyle{IEEEbib}
\bibliography{refs}

\end{document}